\documentclass[runningheads]{llncs}
 
\usepackage[mobile]{eccv}

\usepackage{eccvabbrv}

\usepackage{graphicx}
\usepackage{booktabs}
\usepackage{multirow}
\usepackage{pifont}
\usepackage{caption}

\usepackage[accsupp]{axessibility}  % Improves PDF readability for those with disabilities.

\usepackage[pagebackref,breaklinks,colorlinks,citecolor=eccvblue]{hyperref}
\usepackage{orcidlink}

\begin{document}

% \newcolumntype{L}[1]{>{\raggedright\let\newline\\\arraybackslash\hspace{0pt}}m{#1}}
% \newcolumntype{C}[1]{>{\centering\let\newline\\\arraybackslash\hspace{0pt}}m{#1}}
% \newcolumntype{R}[1]{>{\raggedleft\let\newline\\\arraybackslash\hspace{0pt}}m{#1}} 
\newcommand{\xpar}[1]{\noindent\textbf{#1}\ \ }
\newcommand{\vpar}[1]{\vspace{3mm}\noindent\textbf{#1}\ \ }

\newcommand{\sect}[1]{Section~\ref{#1}}
\newcommand{\sects}[1]{Sections~\ref{#1}}
\newcommand{\eqn}[1]{Equation~\ref{#1}}
\newcommand{\eqns}[1]{Equations~\ref{#1}}
\newcommand{\fig}[1]{Figure~\ref{#1}}
\newcommand{\figs}[1]{Figures~\ref{#1}}
\newcommand{\tab}[1]{Table~\ref{#1}}
\newcommand{\tabs}[1]{Tables~\ref{#1}}
\newcommand{\x}{\mathbf{x}}
\newcommand{\y}{\mathbf{y}}
\newcommand{\fid}{Fr\'echet Inception Distance\xspace}
\newcommand{\lblfig}[1]{\label{fig:#1}}
\newcommand{\lblsec}[1]{\label{sec:#1}}
\newcommand{\lbleq}[1]{\label{eq:#1}}
\newcommand{\lbltbl}[1]{\label{tbl:#1}}
\newcommand{\lblalg}[1]{\label{alg:#1}}
\newcommand{\lblline}[1]{\label{line:#1}}

\newcommand{\ignorethis}[1]{}
\newcommand{\norm}[1]{\lVert#1\rVert_1}
\newcommand{\fcseven}{$\mbox{fc}_7$}

\newsavebox\CBox
\def\textBF#1{\sbox\CBox{#1}\resizebox{\wd\CBox}{\ht\CBox}{\textbf{#1}}}
\renewcommand*{\thefootnote}{\fnsymbol{footnote}}

\def\naive{na\"{\i}ve\xspace}
\def\Naive{Na\"{\i}ve\xspace}

\makeatletter
\DeclareRobustCommand\onedot{\futurelet\@let@token\@onedot}
\def\@onedot{\ifx\@let@token.\else.\null\fi\xspace}

\def\iid{\emph{i.i.d}\onedot}
\def\eg{\emph{e.g}\onedot} \def\Eg{\emph{E.g}\onedot}
\def\ie{\emph{i.e}\onedot} \def\Ie{\emph{I.e}\onedot}
\def\cf{\emph{c.f}\onedot} \def\Cf{\emph{C.f}\onedot}
\def\etc{\emph{etc}\onedot} \def\vs{\emph{vs}\onedot}
\def\wrt{w.r.t\onedot} \def\dof{d.o.f\onedot}
\def\etal{\emph{et al}\onedot}
\def\vs{\textbf{\emph{vs}\onedot}}
\makeatother

\definecolor{citecolor}{RGB}{34,139,34}
\definecolor{mydarkblue}{rgb}{0,0.08,1}
\definecolor{mydarkgreen}{rgb}{0.02,0.6,0.02}
\definecolor{mydarkred}{rgb}{0.8,0.02,0.02}
\definecolor{mydarkorange}{rgb}{0.40,0.2,0.02}
\definecolor{mypurple}{RGB}{111,0,255}
\definecolor{myred}{rgb}{1.0,0.0,0.0}
\definecolor{mygold}{rgb}{0.75,0.6,0.12}
\definecolor{mydarkgray}{rgb}{0.66, 0.66, 0.66}

\newcommand\scalemath[2]{\scalebox{#1}{\mbox{\ensuremath{\displaystyle #2}}}}
\newcommand{\supplement}[1]{\color{blue}{#1}}
\newcommand{\lahav}[1]{{\color{mydarkblue}{[}Lahav: #1{]}}}
\newcommand{\yihan}[1]{{\color{purple}{[}Yihan: #1{]}}}
\newcommand{\JD}[1]{{\color{red}{[}JD: #1{]}}}
\newcommand{\myparagraph}[1]{\paragraph{#1}}

\def\multirowcenter{-0.5\dimexpr \aboverulesep + \belowrulesep + \cmidrulewidth}
\renewcommand{\thefootnote}{\number\value{footnote}}

\newcommand\blfootnotetext[1]{%
  \begingroup
  \renewcommand\thefootnote{}\footnotetext{#1}%
  %\addtocounter{footnote}{-1}%
  \endgroup
}

\title{SEA-RAFT: Simple, Efficient, Accurate RAFT for Optical Flow} 

\author{Yihan Wang\and Lahav Lipson\and Jia Deng}

\authorrunning{Y. Wang et al.}

\institute{Department of Computer Science, Princeton University\\
\email{\{yw7685, llipson, jiadeng\}@princeton.edu}}

\maketitle

\begin{abstract}

We introduce SEA-RAFT, a more simple, efficient, and accurate RAFT for optical flow. Compared with RAFT, SEA-RAFT is trained with a new loss (mixture of Laplace). It directly regresses an initial flow for faster convergence in iterative refinements and introduces rigid-motion pre-training to improve generalization. SEA-RAFT achieves state-of-the-art accuracy on the Spring benchmark with a 3.69 endpoint-error (EPE) and a 0.36 1-pixel outlier rate (1px), representing 22.9\% and 17.8\% error reduction from best published results. In addition, SEA-RAFT obtains the best cross-dataset generalization on KITTI and Spring. With its high efficiency, SEA-RAFT operates at least 2.3$\times$ faster than existing methods while maintaining competitive performance. The code is publicly available at \href{https://github.com/princeton-vl/SEA-RAFT}{https://github.com/princeton-vl/SEA-RAFT}. 

\end{abstract}

\section{Introduction}
\label{sec:intro}

Optical flow is a fundamental task in low-level vision and aims to estimate per-pixel 2D motion between video frames. It is useful for various downstream tasks including action recognition~\cite{sun2018optical, piergiovanni2019representation, zhao2020improved}, video in-painting~\cite{kim2019deep, xu2019deep, gao2020flow}, frame interpolation~\cite{xu2019quadratic, liu2020video, huang2020rife}, 3D reconstruction and synthesis~\cite{ma2022multiview, zuo2022view}.

Although traditionally formulated as an optimization problem~\cite{horn1981determining, zach2007duality, chen2016full}, almost all recent methods are based on deep learning~\cite{teed2020raft, sui2022craft, sun2022skflow, huang2022flowformer, shi2023flowformer++, xu2023unifying, xu2022gmflow, dong2023rethinking, zhao2022global, sun2018pwc, zheng2022dip, zhai2019skflow, weinzaepfel2023croco, leroy2023win, garrepalli2023dift, lu2023transflow, deng2023explicit, weinzaepfel2022croco, saxena2024surprising}. In particular, many state-of-the-art methods~\cite{teed2020raft, sui2022craft, huang2022flowformer, shi2023flowformer++, zhao2022global, lu2023transflow} have adopted architectures based on RAFT~\cite{teed2020raft}, which uses a recurrent network to iteratively refine a flow field. 

\begin{figure}[t]
    \centering
    \includegraphics[width=1.0\linewidth]{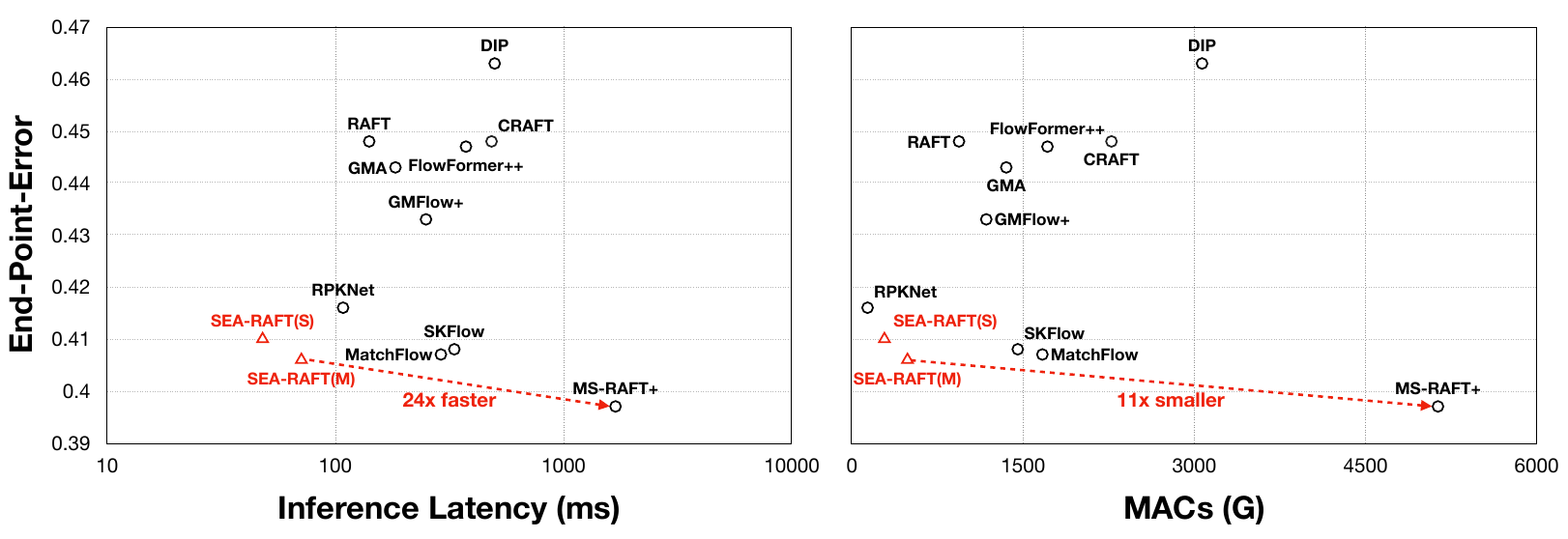}
    \vspace{-0mm}
    \caption{Zero-shot performance of SEA-RAFT and existing methods on the Spring~\cite{mehl2023spring} training split. Latency is measured on an RTX3090 with a batch size of 1 and input resolution $540\times 960$. SEA-RAFT has an accuracy close to the best one achieved by MS-RAFT+~\cite{jahedi2023ms} but is $11\times$ smaller and $24\times$ faster.}
    
    \label{fig:speed}
    \vspace{-0mm}
\end{figure}

In this paper, we introduce SEA-RAFT, a new variant of RAFT that is more efficient and accurate. When compared against all existing approaches, SEA-RAFT has the best accuracy-efficiency Pareto frontier (Fig.~\ref{fig:speed}):
\begin{itemize}
    \item \textit{Accuracy}: On Spring~\cite{mehl2023spring}, SEA-RAFT achieves a new state of the art, outperforming the next best by a large margin: 18\% error reduction on 1px-outlier rate (3.686 vs. 4.482) and 24\% error reduction on endpoint-error (0.363 vs. 0.471). On Sintel~\cite{butler2012naturalistic} and KITTI~\cite{menze2015object}, it outperforms all other methods that have similar computational costs.
    \item \textit{Efficiency}: On each benchmark tested, SEA-RAFT runs at least $2.3\times$ faster than existing methods that have comparable accuracy. Our smallest model, which still outperforms all other methods on Spring, can run at 21fps when processing 1080p images on an RTX3090, 3$\times$ faster than the original RAFT. %\JD{need number for original RAFT}
\end{itemize}

We achieve this by introducing a combination of improvements over the original RAFT: 
\begin{itemize}
    \item \textit{Mixture of Laplace Loss: } Instead of the standard $L_1$ loss, we train the network to predict parameters of a mixture of Laplace distributions to maximize the log-likelihood of the ground truth flow. As we will demonstrate, this new loss reduces overfitting to ambiguous cases and improves generalization. 
    \item \textit{Directly Regressed Initial Flow:} Instead of initializing the flow field to zero before iterative refinement, we directly predict the initial flow by reusing the existing context encoder and feeding it the stacked input frames. This simple change introduces minimal overhead but is surprisingly effective in reducing the number of iterations and improving efficiency.  
    \item \textit{Rigid-Flow Pre-Training:} We find that pre-training on TartanAir~\cite{wang2020tartanair}, which can significantly improve generalization, despite the limited diversity of flow, which is induced purely by camera motion in a static scene.
\end{itemize}

These improvements are novel in the context of RAFT-style methods for optical flow. Moreover, they are orthogonal to the improvements proposed in existing RAFT-style methods, which focus on replacing certain blocks with newer designs, such as replacing convolutional blocks with transformers. 

Besides the main improvements above, SEA-RAFT also incorporates architectural changes that greatly simplify the original RAFT. In particular, we find that certain custom designs of the original RAFT are unnecessary and can be replaced with standard off-the-shelf modules. For example, the original feature encoder and context encoder were custom-designed and must use different normalization layers for stable training; we replaced each with a standard ResNet. In addition, we replace the original convolutional GRU with a simple RNN consisting entirely of ConvNext blocks. Such simplifications make it easy for SEA-RAFT to incorporate new neural building blocks and scale to larger datasets. 

We perform extensive experiments to evaluate SEA-RAFT on standard benchmarks including Spring, Sintel, and KITTI. We also validate the effectiveness of our improvements through ablation studies.

\section{Related Works}

\noindent\textbf{Estimating Optical Flow} Classical approaches treated optical flow as an optimization problem that maximizes visual similarity between corresponding pixels, with strong regularization.\cite{horn1981determining, zach2007duality, chen2016full}. Current methods~\cite{ilg2017flownet, dosovitskiy2015flownet, sun2018pwc, zhao2020maskflownet, hui2018liteflownet, teed2020raft, sui2022craft, sun2022skflow, deng2023explicit, huang2022flowformer, shi2023flowformer++, weinzaepfel2022croco, weinzaepfel2023croco, xu2022gmflow, xu2023unifying, leroy2023win, saxena2024surprising, jahedi2024ccmr, jahedi2023ms, luo2022kpa, zheng2022dip, zhao2022global, luo2023gaflow} are mostly based on deep learning. FlowNets~\cite{dosovitskiy2015flownet, ilg2017flownet} regarded optical flow as a dense regression problem and used stacked convolution blocks for prediction. DCNet~\cite{xu2017accurate} and PWC-Net~\cite{sun2018pwc} introduced 4D cost-volume to explicitly model pixel correspondence. RAFT~\cite{teed2020raft} further combined multi-scale 4D cost-volume with recurrent iterative refinements, achieving large improvements and spawning many follow-ups~\cite{huang2022flowformer, shi2023flowformer++, sun2022skflow, zhao2022global, luo2022kpa, luo2023gaflow, zheng2022dip, jahedi2023ms, sui2022craft}. 

Our method is a new variant of RAFT~\cite{teed2020raft} with several improvements including a new loss function, direct regression of initial flow, rigid-flow pre-training, and architectural simplifications. All of these improvements are new compared to existing RAFT variants. 
In particular, our direct regression of initial flow is new compared to existing efficient RAFT variants~\cite{garrepalli2023dift, morimitsu2024recurrent, deng2023explicit}, which mainly focus on efficient implementations of RAFT modules. This direct regression is a simple change with minimal overhead, but substantially reduces the number of RAFT iterations needed.

\smallskip\noindent\textbf{Data for Optical Flow} FlyingChairs and FlyingThings3D~\cite{dosovitskiy2015flownet, mayer2016large} are commonly used datasets for optical flow. They provide a large amount of synthetic data but have limited realism. Sintel~\cite{butler2012naturalistic}, VIPER~\cite{richter2017playing}, Infinigen~\cite{raistrick2023infinite}, and Spring~\cite{mehl2023spring} are more realistic, using open-source 3D animations, games or procedurally generated scenes. Besides synthetic data, Middlebury, KITTI, and HD1K~\cite{geiger2013vision, kondermann2016hci, menze2015object, baker2011database} provide annotations for real-world image pairs. These datasets are limited in both quantity and diversity due to the difficulty of accurately annotating optical flow in the real world. To leverage more data, several methods~\cite{dong2023rethinking, weinzaepfel2022croco, weinzaepfel2023croco, saxena2024surprising} pre-train their models on different tasks. MatchFlow~\cite{dong2023rethinking} pre-trains on geometric image matching (GIM) using MegaDepth~\cite{li2018megadepth}. Croco-Flow~\cite{weinzaepfel2022croco, weinzaepfel2023croco}, DDVM~\cite{saxena2024surprising}, and Flowformer++~\cite{shi2023flowformer++} pre-train on unlabeled data. We pre-train SEA-RAFT on rigid flow using TartanAir~\cite{wang2020tartanair}. Though TartanAir~\cite{wang2020tartanair} has been used in other methods such as DDVM~\cite{saxena2024surprising} and CroCo-Flow~\cite{weinzaepfel2022croco,weinzaepfel2023croco}, our adoption of rigid-flow pre-training is new in the context of RAFT-style methods.

\smallskip\noindent\textbf{Predicting Probability Distributions} Predicting probability distributions is a common practice in computer vision~\cite{luo2022learning, li2021human, blundell2015weight, truong2023pdc,wannenwetsch2017probflow, sun2021loftr, chen2022aspanformer, zhang2023rgm}. In tasks closely related to optical flow such as keypoint matching~\cite{sun2021loftr, chen2022aspanformer, zhang2023rgm, truong2023pdc}, the variance of the probability distribution reflects uncertainty of predictions and therefore is useful for many applications. For example, LoFTR~\cite{sun2021loftr} filters out uncertain matching pairs. Aspanformer~\cite{chen2022aspanformer} adjusts the look-up radius based on uncertainty. 

To handle the ambiguity caused by heavy occlusion, SEA-RAFT predicts a mixture of Laplace (MoL) distribution. Although MoL has been used in keypoint matching methods such as PDC-Net+~\cite{truong2023pdc}, our use of MoL is new in the context of RAFT-style methods. In addition, our formulation is different in that we require one mixture component to have a constant variance, making it equivalent to the $L_1$ loss that aligns better with the optical flow evaluation metrics. This difference is crucial for achieving competitive performance in optical flow, where every pixel needs accurate correspondence, unlike keypoint matching, where a subset of reliable matches suffices.

\section{Method}

\begin{figure}[t]
    \centering
    %\vspace{-mm}
    \includegraphics[width=1.0\linewidth]{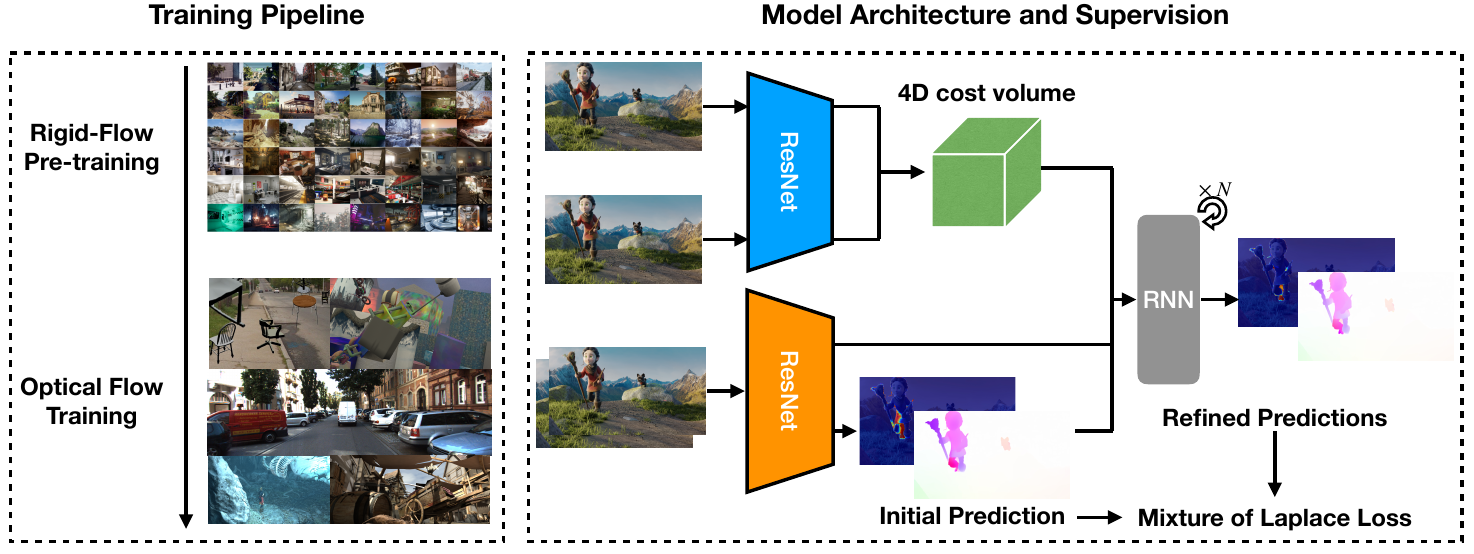}
    \caption{Compared with RAFT~\cite{teed2020raft}, SEA-RAFT introduces (1) rigid-flow pre-training, (2) mixture of Laplace loss, and (3) direct regression of initial flow.}
    \label{fig:pipeline}
    %\vspace{-6mm}
\end{figure}

In this section, we first describe the iterative refinement in RAFT and then introduce the improvements that lead to SEA-RAFT. 

\subsection{Iterative refinement}

Given two adjacent RGB frames, RAFT predicts a field of pixel-wise 2D vectors through iterative refinement that consists of two parts: (1) feature and context encoders, which transform images into lower-resolution dense features, and (2) an RNN unit, which iteratively refines the predictions.

Given two images $I_{1}, I_{2}\in \mathbb{R}^{H\times W \times 3}$, the feature encoder $F$ takes $I_1$, $I_2$ as inputs separately and outputs a lower-resolution feature $F(I_1), F(I_2)\in \mathbb{R}^{h\times w \times D}$. The context encoder $C$ takes source image $I_1$ as input and outputs a context feature $C(I_1)\in  \mathbb{R}^{h\times w \times D}$. A multi-scale 4D correlation volume $\{V_k\}$ is then built with the features from feature encoder $F$:
\begin{align*}
V_k = F(I_1) \circ \texttt{AvgPool}(F(I_2), 2^k)^{\top}\in\mathbb{R}^{h\times w\times \frac{h}{2^k}\times\frac{w}{2^k}},
\end{align*}
where $\circ$ represents the correlation operator, which computes similarities (as dot products of feature vectors) between all pairs of pixels across two feature maps.

Several works~\cite{jahedi2023ms,jahedi2024ccmr} have explored the optimal choices of the number of levels in the cost volume($k$) and the feature resolution $(h, w)$. In SEA-RAFT, we simply follow the original setting in RAFT~\cite{teed2020raft}: $(h, w) = \frac{1}{8}(H, W), k=4$.

RAFT iteratively refines a flow prediction $\mu$. Initially, $\mu$ is set to be all zeros. Each refinement step uses the current flow prediction $\mu$ to fetch a $D_M$-dim motion feature $M$ from the multi-scale correlation volume $\{V_k\}$ with a look-up radius $r$:
\begin{align*}
    M = \texttt{MotionEncoder}(\texttt{LookUp}(\{V_k\}, \mu, r)) \in \mathbb{R}^{h\times w\times D_M},
\end{align*}
where the $\texttt{Lookup}$ operator returns a motion feature vector for each pixel in $I_1$, consisting of similarities between the pixel in $I_1$ and its current correspondence's neighboring pixels in $I_2$ within the radius $r$. The motion feature vector is further transformed by a motion encoder. 

Existing works~\cite{chen2022aspanformer, garrepalli2023dift, jung2023anyflow} have explored dynamic radius and look-up when obtaining the motion features from $\{V_k\}$. For simplicity of design, SEA-RAFT follows the original RAFT and sets the look-up radius $r=4$ to a fixed constant. The motion feature $M$ is fed into the RNN cell along with hidden state $h$ and context feature $C(I_1)$. From the new hidden state $h'$, the residual flow $\Delta\mu$ is regressed by a 2-layer $\texttt{FlowHead}$:
\begin{align*}
 h' &= \texttt{RNN}(h, M, C(I_1))\\
 \Delta \mu &= \texttt{FlowHead}(h')
\end{align*}

Methods using RAFT-Style iterative refinement~\cite{teed2020raft, huang2022flowformer} usually need many iterations: $12$ in training and as many as $32$ in inference. As a result, RNN-based iterative refinement is a significant bottleneck in latency. Though there have been attempts~\cite{deng2023explicit, garrepalli2023dift} to reduce the number of iterations, the performance drastically drops with fewer iterations. In contrast, SEA-RAFT only needs $4$ iterations in training and up to $12$ iterations in inference to achieve competitive performance.  

\subsection{Mixture-of-Laplace Loss}

\begin{figure}[t]
    \centering
    \includegraphics[width=1.0\linewidth]{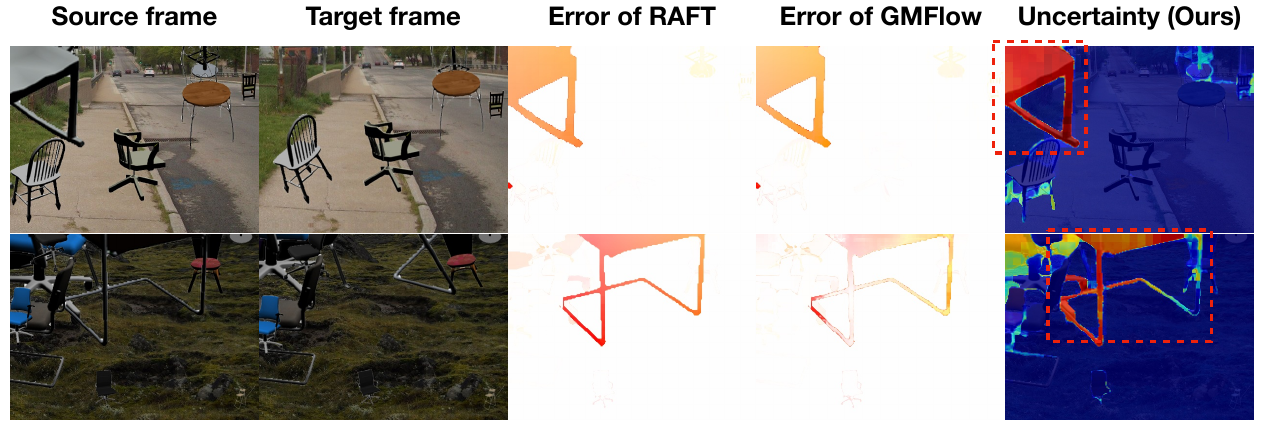}
    \vspace{-0mm}
    \caption{Ambiguous cases can occur frequently in training data where flow is unpredictable due to occlusion. Such cases can dominate the $L_1$ loss (shown as an error map) used by current methods~\cite{teed2020raft, xu2022gmflow}. Our new training loss allows the model to account for such uncertainty.}
    \label{fig:ambi}
    \vspace{-0mm}
\end{figure}

 Most prior works are supervised using an endpoint-error loss on all pixels. However, optical flow training data often contains ambiguous, unpredictable samples, which can dominate this loss empirically.

\smallskip\noindent\textbf{Ambiguous Cases} Ambiguous cases of optical flow can arise with heavy occlusion~\cref{fig:ambi}. While in many cases the motion of occluded pixels can be predicted, sometimes the ambiguity can be too large to predict a single outcome. We examined 10 samples with the highest endpoint-error in the training and validation sets of FlyingChairs~\cite{dosovitskiy2015flownet} and found that ambiguous cases dominate the error.

\label{sec:prob}
\smallskip\noindent\textbf{Review of Probabilistic Regression} 
Prior works for image-matching have proposed probabilistic losses to enable their model to express aleatoric or epistemic uncertainty~\cite{sun2021loftr, truong2023pdc, chen2022aspanformer,truong2023pdc, weinzaepfel2023croco, wannenwetsch2017probflow, zhang2023rgm}. These approaches regress the parameters of the probabilistic model and maximize the log-likelihood of the ground truth during training.

\noindent Given an image pair $\{I_{1},I_{2}\}$ and the flow ground truth $\mu_{gt}$, the training loss is\begin{align*}
    \mathcal{L}_{prob} = -\log{p_{\theta}(\mu=\mu_{gt}|I_{1}, I_{2})}
\end{align*}
where the probability density function $p_\theta$ is parameterized by the network. Prior work has formulated $p_{\theta}$ as a Gaussian or a Laplace distribution with a predicted mean and variance. For example, we can formulate a naive version of probabilistic regression by assuming: (1) $p_{\theta}$ is Laplace with mean $\mu\in\mathbb{R}^{H\times W\times 2}$ and scale $b\in\mathbb{R}^{H\times W\times 1}$ predicted by the network, (2) the flow distribution is pixel-wisely independent, and (3) the x-direction flow and the y-direction flow are independent but share the same scale parameter $b$:
\begin{align}
    \mathcal{L}_{Lap} &= \frac{1}{HW}\sum_{u}\sum_{v}{(\log{2b(u, v)} + \frac{\|\mu_{gt}(u, v) - \mu(u, v)\|_1}{2b(u, v)})}
\end{align}
where $u,v$ are indices to the pixels. 
The Laplace loss can be regarded as an extended version of $L_1$ loss with an extra penalty term $b$. During inference, $\mu$ represents the flow prediction, and the scale factor $b$ provides an estimation of uncertainty. However, we find this naive probabilistic regression does not work well on optical flow, which has also been pointed out by prior work~\cite{zhang2023rgm}.

\begin{figure}[t]
    \centering
    \includegraphics[width=1.0\linewidth]{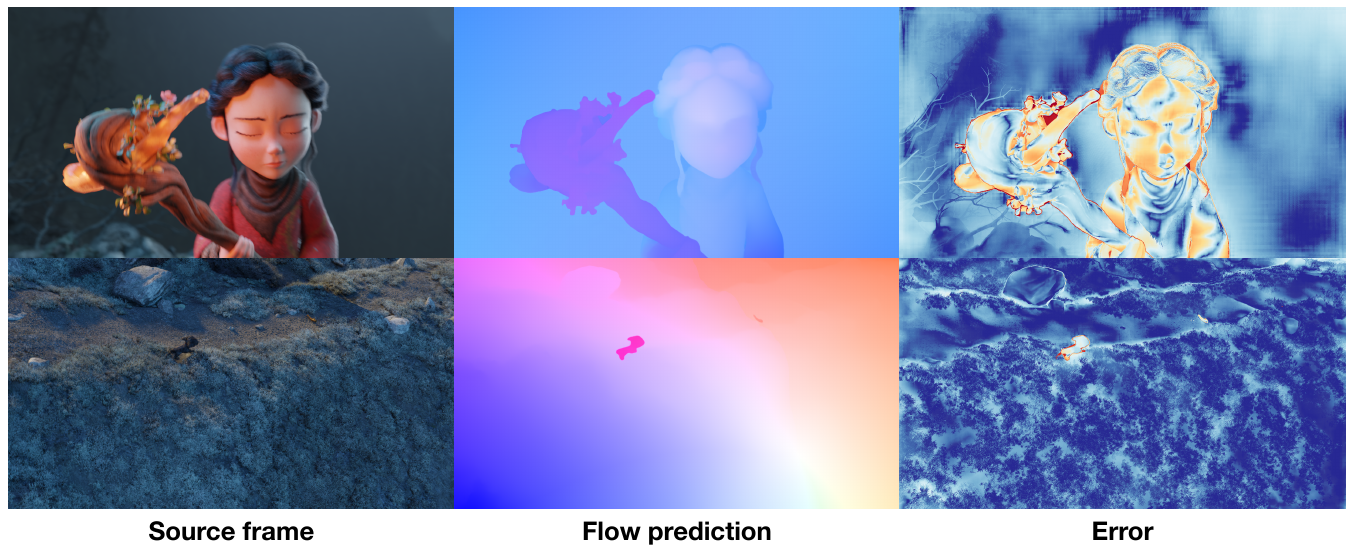}
    % \vspace{-4mm}
    \caption{Visualization on Spring~\cite{mehl2023spring} test set.}
    \label{fig:vis-spring}
    % \vspace{-4mm}
\end{figure}

\smallskip\noindent\textbf{Mixture of Laplace} One reason that naive probabilistic regression performs poorly is numerical instability as the loss contains a log term. To address this issue, we regress $b(u,v)$ directly in log-space. This approach makes training more stable compared to previous approaches which clamp $b$ to $[\epsilon,\infty)$, where $\epsilon$ is a small positive number. 

Another reason that naive probabilistic regression performs poorly is that it deviates from the standard endpoint-error metric, which only cares about the $L_1$ difference, but not the uncertainty estimation. Thus, we propose to use a mixture of two Laplace distributions: one for ordinary cases, and the other for ambiguous cases, with mixing coefficient $\alpha \in [0,1]$:
\begin{align*}
    MixLap(x;\alpha, \beta_1, \beta_2, \mu) &= \alpha\cdot\frac{e^{-\frac{|x-\mu|}{e^{\beta_1}}}}{2e^{\beta_1}} + (1-\alpha)\cdot\frac{e^{-\frac{|x-\mu|}{e^{\beta_2}}}}{2e^{\beta_2}}
\end{align*}

 Intuitively, at each pixel, we want the first component of the mixture to be aligned with the endpoint-error metric, and the second component to account for ambiguous cases. To explicitly enforce this, we fix $\beta_1=0$, such that the network is encouraged to optimize for the L1 loss when possible. This leads to the following Mixture-of-Laplace (MoL) loss:
\begin{align}
    \mathcal{L}_{MoL} &= \scalemath{0.85}{-\frac{1}{2HW}\sum_{u}\sum_{v}\sum_{d\in \{x,y\}}{\log{[MixLap(\mu_{gt}(u,v)_d;\alpha(u,v), 0, \beta_2(u,v), \mu(u,v)_d)]}}}
\end{align}
where $d$ indexes the axe of the flow vector (the $x$ direction or $y$ direction).

The free parameters $\alpha$, $\beta_2$, $\mu$ of $\mathcal{L}_{MoL}$ are predicted by the network. 
Intuitively, a higher $\alpha$ means the flow prediction of this pixel is more ``ordinary'' instead of ``ambiguous''. Mathematically, a higher $\alpha$ makes $\mathcal{L}_{moL}$ behave like an $L_1$ loss. In~\cref{sec:ablation}, we then show that this property leads to better accuracy. 
    
Note that though the mixture model has been used in keypoint matching~\cite{sun2021loftr, chen2022aspanformer, truong2023pdc}, its application to optical flow requires a different formulation because the goal is substantially different. In keypoint matching, the goal is to identify a \emph{subset} of reliable matches for downstream applications such as camera pose estimation. Predicting uncertainty serves to filter out unreliable matches, and there is no explicit penalty for predicting few correspondences. As a result, it is not essential for them to align a mixing component to $L_1$ loss. In optical flow, we are evaluated on the flow prediction for \emph{every} pixel.

\smallskip\noindent\textbf{Implementation Details} We set an upper bound for $\beta$ to $10$ in the loss to make the training more stable. 
We also re-predict $\alpha$ and $\beta$ every update iteration. We can similarly define the probabilistic sequence loss as:\vspace{-0mm}% I added this because there was a ton of whitespace -Lahav
\begin{align}
    \mathcal{L}_{all} = \sum_{i=1}^{N}{\gamma^{N-i}\mathcal{L}_{MoL}^{i}}
\end{align}
where $\mathcal{L}_{mix}^{i}$ denotes the probabilistic loss in iteration $i$, $N$ denotes the number of iterations, and $\gamma < 1$ exponentially downweights the early iterations. We empirically observe that our method significantly reduces the number of update iterations needed in inference. In fact, $N=4$ is sufficient for SEA-RAFT to take first place on the Spring~\cite{mehl2023spring} benchmark. We provide detailed ablations in~\cref{tab:ablations}.

\subsection{Direct Regression of Initial Flow}
\label{sec: initflow}

\begin{figure}[t]
    \centering
    \includegraphics[width=1.0\linewidth]{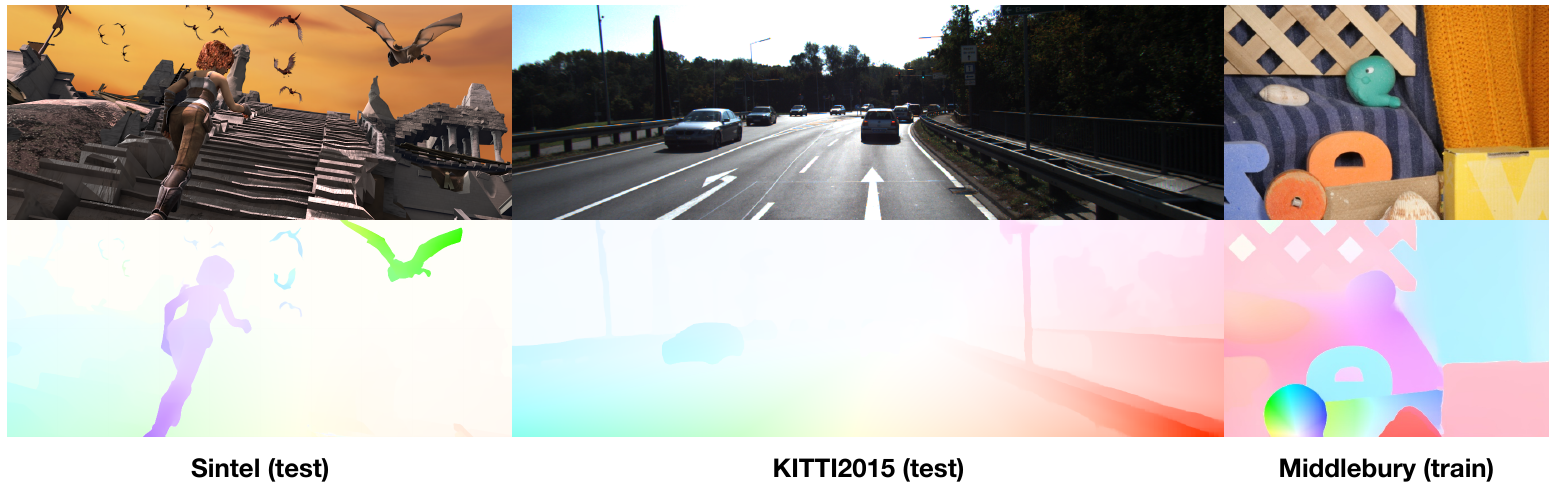}
    \caption{Visualization on Sintel~\cite{butler2012naturalistic}, KITTI~\cite{menze2015object}, and Middlebury~\cite{baker2011database}.}
    \label{fig:vis-other}
    %\vspace{-4mm}
\end{figure}

RAFT-style iterative refinements~\cite{sun2022skflow, huang2022flowformer, morimitsu2024recurrent, dong2023rethinking, luo2023gaflow, zhao2022global} typically zero-initialize the flow field. However, zero-initialization may deviate substantially from the ground truth, thus needing many iterations. In SEA-RAFT, we borrow an idea from the FlowNet family of methods~\cite{ilg2017flownet, dosovitskiy2015flownet} to predict an initial estimate of optical flow from the context encoder, given both frames as input. We also predict an associated MoL (see ~\cref{sec:prob}).

This simple modification also significantly improves the convergence speed of the iterative refinement framework, allowing one to use fewer iterations during inference. Detailed ablations are shown in~\cref{tab:ablations}.

\subsection{Large-Scale Rigid-Flow Pre-Training}
\label{sec: stereo}

Most prior works train on a small number of datasets with limited size, diversity and realism~\cite{dosovitskiy2015flownet, mayer2016large}. To improve generationalization, we pre-train SEA-RAFT on TartanAir~\cite{wang2020tartanair}, which provides optical flow annotations between a pair of (non-rectified) stereo cameras. This type of motion field is a special case of optical flow due to viewpoint change in a rigid static scene. Despite its limited motion diversity, it enables SEA-RAFT to train on data with higher realism and scene diversity, leading to better generalization. \looseness=-1

\subsection{Simplifications}

We also provide a few architecture changes that greatly simplify the original RAFT~\cite{teed2020raft}. First, we adopt truncated, ImageNet~\cite{deng2009imagenet} pre-trained ResNets for the backbones. We also substitute the ConvGRU in RAFT with two ConvNeXt~\cite{liu2022convnet} blocks, which we show provides better efficiency and training stability. The detailed ablations of these changes are shown in~\cref{tab:ablations}.

\section{Experiments}

We evaluate SEA-RAFT on Spring~\cite{mehl2023spring}, KITTI~\cite{geiger2013vision}, and Sintel~\cite{butler2012naturalistic}. Following previous works, we also incorporate FlyingChairs~\cite{dosovitskiy2015flownet}, FlyingThings~\cite{mayer2016large}, and HD1K~\cite{kondermann2016hci} into our training pipeline. To verify the effectiveness of TartanAir~\cite{wang2020tartanair} rigid-flow pre-training, we provide the performance gain from it in different settings.

\smallskip\noindent\textbf{Model Details} SEA-RAFT is implemented in PyTorch~\cite{paszke2019pytorch}. There are three different types of SEA-RAFT and we denote them as \textBF{SEA-RAFT(S/M/L)}. The only differences among them are the backbone choices and the number of iterations in inference. Specifically, SEA-RAFT(S) uses the first 6 layers of ResNet-18 as the feature/context encoder, and SEA-RAFT(M) uses the first 13 layers of ResNet-34. The pre-trained weights we use are downloaded from torchvision. SEA-RAFT(S) and SEA-RAFT(M) use the same architecture for the recurrent units and keep the number of iterations $N=4$ in both training and inference. SEA-RAFT(L) can be regarded as an extension based on SEA-RAFT(M): they share the same weights, but SEA-RAFT(L) uses $N=12$ iterations in inference. Following RAFT~\cite{teed2020raft}, we stop the gradient for $\mu$ when computing $\mu'=\mu + \Delta \mu$ and only propagate the gradient for residual flow $\Delta \mu$.

\smallskip\noindent\textbf{Training Details} As mentioned in~\cref{sec: stereo}, We pre-train SEA-RAFT on TartanAir~\cite{wang2020tartanair} for 300k steps with a batch size of 32, input resolution $480\times 640$ and learning rate $4\times 10^{-4}$. Similar to RAFT~\cite{teed2020raft}, MaskFlowNet~\cite{zhao2020maskflownet} and PWC-Net+~\cite{sun2018pwc}, we then train our models on FlyingChairs~\cite{dosovitskiy2015flownet} for 100k steps with a batch size of 16, input resolution $368\times 496$, learning rate $2.5 \times 10^{-4}$ and FlyingThings3D~\cite{mayer2016large} for 120k steps with a batch size of 32, input resolution $432\times 960$, learning rate $4\times 10^{-4}$ (denoted as "C+T" following previous works). For the submissions on Sintel~\cite{butler2012naturalistic} benchmark, we fine-tune the model from "C+T" on a mixture of Sintel~\cite{butler2012naturalistic}, FlyingThings3D clean pass~\cite{mayer2016large}, KITTI~\cite{geiger2013vision} and HD1K~\cite{kondermann2016hci} for 300k steps with a batch size of 32, input resolution $432\times 960$ and learning rate $4\times 10^{-4}$ (denoted as "C+T+S+K+H" following previous works). Different from previous methods, we reduce the percentage of Sintel~\cite{butler2012naturalistic} in the mixture dataset, which is usually more than 70\% in previous papers. Details will be mentioned in the supplementary material. For KITTI~\cite{geiger2013vision} submissions, we fine-tune our models from "C+T+S+K+H" on the KITTI training set for extra 10k steps with a batch size of 16, input resolution $432\times 960$ and learning rate $10^{-4}$. For Spring~\cite{mehl2023spring} submissions, we fine-tune our models from "C+T+S+K+H" on the Spring training set for extra 120k steps with a batch size of 32, input resolution $540\times 960$ and learning rate $4\times 10^{-4}$.

\smallskip\noindent\textbf{Metrics} We adopt the widely used metrics in this study: endpoint-error (EPE), 1-pixel outlier rate (1px), Fl-score and WAUC error. Definitions can be found in~\cite{richter2017playing, mehl2023spring, geiger2013vision}.

\setlength\tabcolsep{.2em}
\begin{table*}[t]
    \centering
    \resizebox{1.0\linewidth}{!}{
    \begin{tabular}{llccccccc}
    \toprule
    \multirow{2}{*}[\multirowcenter]{Extra Data} & \multirow{2}{*}[\multirowcenter]{Method} &\multicolumn{3}{c}{Spring (train)} &\multicolumn{4}{c}{Spring (test)} \\
    \cmidrule(l{0.5ex}r{0.5ex}){3-5} \cmidrule(l{0.5ex}r{0.5ex}){6-9}
    & & Fine-tune & 1px$\downarrow$ & EPE$\downarrow$ & 1px$\downarrow$ & EPE$\downarrow$ & Fl$\downarrow$ & WAUC$\uparrow$ \\ 
    \midrule
        & PWC-Net~\cite{sun2018pwc} & $\text{\ding{55}}$ & - & - & 82.27$^*$ & 2.288$^*$ & 4.889$^*$ & 45.670$^*$\\
        & FlowNet2~\cite{ilg2017flownet} & $\text{\ding{55}}$ & - & - & 6.710$^*$ & 1.040$^*$ & 2.823$^*$ & 90.907$^*$\\
        & RAFT~\cite{teed2020raft} & $\text{\ding{55}}$ & 4.788 & 0.448 & 6.790$^*$ & 1.476$^*$ & 3.198$^*$ & 90.920$^*$\\
        & GMA~\cite{jiang2021learning} & $\text{\ding{55}}$ & 4.763 & 0.443 & 7.074$^*$ & 0.914$^*$ & 3.079$^*$ & 90.722$^*$\\
        & RPKNet~\cite{morimitsu2024recurrent} & $\text{\ding{55}}$ & 4.472 & 0.416 & 4.809 & 0.657 & 1.756 & 92.638\\
        & DIP~\cite{zheng2022dip} & $\text{\ding{55}}$ & 4.273 & 0.463 & - & - & - & -\\
        & SKFlow~\cite{sun2022skflow} & $\text{\ding{55}}$ & 4.521 & 0.408 & - & - & - & -\\
        & GMFlow~\cite{xu2022gmflow} & $\text{\ding{55}}$ & 29.49 & 0.930 & 10.355$^*$ & 0.945$^*$ & 2.952$^*$ & 82.337$^*$\\
        & GMFlow$+$~\cite{xu2023unifying} & $\text{\ding{55}}$ & 4.292 & 0.433 & - & - & - & -\\
        & Flowformer~\cite{huang2022flowformer} & $\text{\ding{55}}$ & 4.508 & 0.470 & 6.510$^*$ & 0.723$^*$ & 2.384$^*$ & 91.679$^*$\\
        & CRAFT~\cite{sui2022craft} & $\text{\ding{55}}$ & 4.803 & 0.448 & - & - & - & -\\
        & \textBF{SEA-RAFT(S)} & $\text{\ding{55}}$ & \underline{4.077} & \underline{0.415} & - & - & - & -\\
        & \textBF{SEA-RAFT(M)} & $\text{\ding{55}}$ & \textBF{4.060} & \textBF{0.406} & - & - & - & -\\
    \midrule
        MegaDepth~\cite{li2018megadepth} & MatchFlow(G)~\cite{dong2023rethinking} & $\text{\ding{55}}$ & 4.504 & 0.407 & - & - & - & -\\
        YouTube-VOS~\cite{xu2018youtube}& Flowformer$++$\cite{shi2023flowformer++}& $\text{\ding{55}}$ & 4.482 & 0.447& - & - & - & - \\ 
        VIPER~\cite{richter2017playing}& MS-RAFT+~\cite{jahedi2023ms} & $\text{\ding{55}}$ & \textBF{3.577} & \textBF{0.397} & 5.724$^*$ & 0.643$^*$ & 2.189$^*$ & 92.888$^*$\\
        TartanAir~\cite{wang2020tartanair}& \textBF{SEA-RAFT(S)} & $\text{\ding{55}}$ & 4.161 & 0.410 & - & - & - & -\\
        TartanAir~\cite{wang2020tartanair}& \textBF{SEA-RAFT(M)} & $\text{\ding{55}}$ & \underline{3.888} & \underline{0.406} & - & - & - & -\\
    \midrule
        CroCo-Pretrain & CroCoFlow~\cite{weinzaepfel2023croco} & $\text{\ding{51}}$ & - & - & 4.565 & 0.498 & 1.508 & 93.660\\
        CroCo-Pretrain & Win-Win~\cite{leroy2023win} & $\text{\ding{51}}$ & - & - & 5.371 & 0.475 & 1.621 & 92.270\\
        TartanAir~\cite{wang2020tartanair}& \textBF{SEA-RAFT(S)} &$\text{\ding{51}}$ & - & - & \underline{3.904} & \underline{0.377} & \underline{1.389} & \underline{94.182}\\
        TartanAir~\cite{wang2020tartanair}& \textBF{SEA-RAFT(M)} &$\text{\ding{51}}$ & - & - & \textBF{3.686} & \textBF{0.363} & \textBF{1.347} & \textBF{94.534}\\
    \bottomrule
    \end{tabular}
    }
    \caption{SEA-RAFT outperforms existing methods on Spring~\cite{mehl2023spring} in different settings. $^*$ denotes the results submitted by Spring~\cite{mehl2023spring} team. By default, all methods have undergone "C+T+S+K+H" training. We list the data used by each method beyond default in the "Extra Data" column. On Spring(test), even our smallest model SEA-RAFT(S) surpasses existing methods by a significant margin. Without fine-tuning on Spring(train), SEA-RAFT outperforms all other methods that do not use extra data.}
    \label{tab:spring}
    % \vspace{-5mm} % Feel free to remove -Lahav
\end{table*}

\subsection{Results on Spring} 
\label{sec:exp-spring}

\smallskip\noindent\textbf{Zero-Shot Evaluation} We compare several representative existing methods with SEA-RAFT using the checkpoints and configurations for Sintel~\cite{butler2012naturalistic} submission on the Spring~\cite{mehl2023spring} training split. For fair comparisons, we remove the test-time optimizations such as tiling in this setting, which will significantly slow down the inference speed. All experiments follow the same downsample-upsample protocol: We first downsample the 1080p images by $2\times$, do inference, and then bi-linearly upsample the flow field back to 1080p, which ensures the input resolution in inference is similar to their training resolution in "C+T+S+K+H". As shown in~\cref{tab:spring}, SEA-RAFT achieves the best results among representative existing methods without using extra data, which demonstrates the superiority of our mixture loss and architecture design. When allowed to use extra data, SEA-RAFT falls slightly behind MS-RAFT+~\cite{jahedi2023ms} but is $24\times$ faster and $11\times$ smaller as mentioned in~\cref{fig:speed}.

\smallskip\noindent\textbf{Fine-Tuning Test} SEA-RAFT ranks 1st on the public test benchmark: SEA-RAFT(M) outperforms all other methods by at least $22.9\%$ on average EPE(endpoint-error) and $17.8\%$ on 1px (1-pixel outlier rate), and SEA-RAFT(S) outperforms other methods by at least $20.0\%$ on EPE and $12.8\%$ on 1px. Besides the strong performance, our method is notably fast. SEA-RAFT(S) is at least $2.3\times$ faster than existing methods which can achieve similar performance. As we still follow the downsample-upsample protocol without using any test-time optimizations in submissions, the inference latency directly reflects our speed in handling 1080p images, which means over 20fps on a single RTX3090.

\subsection{Results on Sintel and KITTI}

\setlength\tabcolsep{.6em}
\begin{table*}[t]
    \centering
    \resizebox{1.0\linewidth}{!}{
    \begin{tabular}{lccccc}
    \toprule
    \multirow{2}{*}[\multirowcenter]{Extra Data} & \multirow{2}{*}[\multirowcenter]{Method} & \multicolumn{2}{c}{Sintel} & \multicolumn{2}{c}{KITTI}\\
    \cmidrule(l{0.5ex}r{0.5ex}){3-4}\cmidrule(l{0.5ex}r{0.5ex}){5-6}
        &  & Clean$\downarrow$ & Final$\downarrow$ & Fl-epe$\downarrow$ & Fl-all$\downarrow$ \\ 
    \midrule
        & PWC-Net~\cite{sun2018pwc} & 2.55 & 3.93 & 10.4 & 33.7\\
        & RAFT~\cite{teed2020raft} & 1.43 & 2.71 & 5.04 & 17.4\\
        & GMA~\cite{jiang2021learning} & 1.30 & 2.74 & 4.69 & 17.1\\
        & SKFlow~\cite{sun2022skflow} & 1.22 & 2.46 & 4.27 & 15.5\\
        & FlowFormer~\cite{huang2022flowformer} & 1.01 & \textBF{2.40} & 4.09$^{\dagger}$ & 14.7$^{\dagger}$\\
        & DIP~\cite{zheng2022dip} & 1.30 & 2.82 & 4.29 & 13.7\\
        & EMD-L~\cite{deng2023explicit} & \textBF{0.88} & 2.55 & 4.12 & 13.5\\
        & CRAFT~\cite{sui2022craft} & 1.27 & 2.79 & 4.88 & 17.5\\
        & RPKNet~\cite{morimitsu2024recurrent} & 1.12 & 2.45 & - & 13.0\\
        & GMFlowNet~\cite{zhao2022global} & 1.14 & 2.71 & 4.24 & 15.4\\
        & \textBF{SEA-RAFT(M)} &1.21 & 4.04 & 4.29 & 14.2\\
        & \textBF{SEA-RAFT(L)} &1.19 & 4.11 & \textBF{3.62} & \textBF{12.9}\\
        \midrule
        & GMFlow~\cite{xu2022gmflow} & 1.08 & 2.48 & 11.2$^{*}$ & 28.7$^{*}$\\
        Tartan & GMFlow~\cite{xu2022gmflow} & - & - & 8.70 (-22\%)$^{*}$ & 24.4 (-15\%)$^{*}$\\
        \midrule
        & \textBF{SEA-RAFT(S)} &1.27 & 4.32 & 4.61 & 15.8\\
        Tartan & \textBF{SEA-RAFT(S)} &1.27 & 3.74 (-13\%) & 4.43 & 15.1\\
        K+H & \textBF{SEA-RAFT(S)} & 1.32 & 2.95 (-32\%)& - & - \\ 
        Tartan+K+H & \textBF{SEA-RAFT(S)} & 1.30 & 2.79 (-35\%) & - & -\\
    \bottomrule
    \end{tabular}
    }
    \caption{SEA-RAFT achieves the best zero-shot performance on KITTI(train). By default, all methods are trained with "C+T". We list the extra data in the first column. $^{\dagger}$ denotes the method uses tiling in inference. $^{*}$ denotes the GMFlow~\cite{xu2022gmflow} ablation with 200k training steps.}
    \label{tab:SK-train}
    % \vspace{-7mm}
\end{table*}

\setlength\tabcolsep{.4em}
\begin{table*}[t]
    \centering
    \resizebox{1.0\linewidth}{!}{
    \begin{tabular}{llccccccc}
    \toprule
    \multirow{2}{*}[\multirowcenter]{Extra Data} & \multirow{2}{*}[\multirowcenter]{Method} & \multicolumn{2}{c}{Sintel} & \multicolumn{3}{c}{KITTI} & \multicolumn{2}{c}{Inference Cost}\\
    \cmidrule(l{0.5ex}r{0.5ex}){3-4}\cmidrule(l{0.5ex}r{0.5ex}){5-7}\cmidrule(l{0.5ex}r{0.5ex}){8-9}
        &  & Clean$\downarrow$ & Final$\downarrow$ & Fl-all$\downarrow$ & Fl-bg$\downarrow$ & Fl-fg$\downarrow$ & \#MACs & Latency\\ 
    \midrule
        & PWC-Net+~\cite{sun2019models} & 3.45 & 4.60 & 7.72 & 7.69 & 7.88 & \textBF{101.3G} & \textBF{23.82ms}\\
        & RAFT~\cite{teed2020raft} & 1.61$^{\star}$ & 2.86$^{\star}$ & 5.10 & 4.74& 6.87& 938.2G & 140.7ms\\
        & GMA~\cite{jiang2021learning} & 1.39$^{\star}$ & 2.47$^{\star}$ & 5.15 & - & - & 1352G & 183.3ms\\
        & DIP~\cite{zheng2022dip} &1.44$^{\star}$ & 2.83$^{\star}$ & 4.21 & 3.86 & 5.96 & 3068G & 498.9ms\\
        & GMFlowNet~\cite{zhao2022global} & 1.39 & 2.65 & 4.79 & 4.39 & 6.84 & 1094G & 244.3ms\\
        & GMFlow~\cite{xu2022gmflow} & 1.74 & 2.90 & 9.32 & 9.67 & 7.57 & 602.6G & 138.5ms\\
        & CRAFT~\cite{sui2022craft} & 1.45$^{\star}$ & 2.42$^{\star}$ & 4.79 & 4.58 & 5.85 & 2274G & 483.4ms\\
        & FlowFormer~\cite{huang2022flowformer} & 1.20 & 2.12 & 4.68$^{\dagger}$ & 4.37$^{\dagger}$ & 6.18$^{\dagger}$ & 1715G & 335.6ms\\
        & SKFlow~\cite{sun2022skflow} & 1.28$^{\star}$ & 2.23$^{\star}$ & 4.85 & 4.55 & 6.39 & 1453G & 331.9ms\\
        & GMFlow+~\cite{xu2023unifying} & \textBF{1.03} & 2.37 & 4.49 & 4.27 & 5.60 & 1177G & 249.6ms\\
        & EMD-L~\cite{deng2023explicit} & 1.32 & 2.51 & 4.49 & 4.16 & 6.15 & 1755G & OOM\\
        & RPKNet~\cite{morimitsu2024recurrent} & 1.31 & 2.65 & 4.64 & 4.63 & \textBF{4.69} & \underline{137.0G} & 183.3ms \\ 
        \midrule
        VIPER~\cite{richter2017playing} & CCMR+~\cite{jahedi2024ccmr} & \underline{1.07} & \underline{2.10} & 3.86 & 3.39 & 6.21 & 12653G & OOM \\ 
        MegaDepth~\cite{li2018megadepth} & MatchFlow(G)~\cite{dong2023rethinking} & 1.16$^{\star}$ & {2.37}$^{\star}$ & {4.63}$^{\dagger}$ & 4.33$^{\dagger}$ & 6.11$^{\dagger}$ & 1669G & 290.6ms\\
        YouTube-VOS~\cite{xu2018youtube}& Flowformer++\cite{shi2023flowformer++} & \underline{1.07} & \textBF{1.94} & 4.52$^{\dagger}$ & - & - & 1713G & 373.4ms\\ 
        CroCo-Pretrain & CroCoFlow~\cite{weinzaepfel2023croco} & 1.09$^{\dagger}$ & 2.44$^{\dagger}$ & \underline{3.64}$^{\dagger}$ & \underline{3.18}$^{\dagger}$ & 5.94$^{\dagger}$ & 57343G$^{\dagger}$ &  6422ms$^{\dagger}$ \\
        DDVM-Pretrain & DDVM~\cite{saxena2024surprising} &1.75$^{\dagger}$ &2.48$^{\dagger}$ & \textBF{3.26}$^{\dagger}$ &  \textBF{2.90}$^{\dagger}$ & \underline{5.05}$^{\dagger}$ & -& -\\
        TartanAir~\cite{wang2020tartanair}& \textBF{SEA-RAFT(M)} &1.44 & 2.86 & 4.64 & 4.47 & 5.49 & 486.9G & \underline{70.96ms}\\
        TartanAir~\cite{wang2020tartanair}& \textBF{SEA-RAFT(L)} &1.31 & 2.60 & 4.30 & 4.08 & 5.37 & 655.1G & 108.0ms\\
    \bottomrule
    \end{tabular}
    }
    \caption{Compared with other methods that achieve competitive performance, SEA-RAFT is at least $1.8\times$ faster on Sintel(test)~\cite{butler2012naturalistic} and $4.6\times$ faster on KITTI(test)~\cite{menze2015object}. All methods have undergone "C+T+S+K+H" training by default and we list the extra data each method uses in the first column. We measure latency on an RTX3090 with a batch size of 1 and input resolution $540\times 960$. $^{\star}$ denotes the method uses warm-start~\cite{teed2020raft} strategy. $^{\dagger}$ denotes that the corresponding methods use tiling-based test-time optimizations.}
    % \vspace{-6mm}
    \label{tab:SK-test}
\end{table*}

\smallskip\noindent\textbf{Zero-Shot Evaluation} Following previous works, we evaluate the zero-shot performance of SEA-RAFT given training schedule "C+T" on Sintel(train)~\cite{butler2012naturalistic} and KITTI(train)~\cite{menze2015object}. The results are provided in~\cref{tab:SK-train}. On KITTI(train), SEA-RAFT outperforms all prior works by a large margin, improving Fl-epe from $4.09$ to $3.62$ and Fl-all from $13.7$ to $12.9$. On Sintel(train), SEA-RAFT achieves competitive results on the clean pass but, for reasons unclear to us, underperforms existing methods on the final pass. Note that although this ``C+T''  zero-shot setting is standard, it is of limited relevance to real-world applications, which do not need to restrict the training data to only C+T. Indeed, we show that by adding a small amount of high-quality real-world data (KITTI + HD1K, about 1.2k image pairs compared with 80k image pairs in FlyingThings3D~\cite{mayer2016large}), the performance gap on the Sintel(train) final pass can be remarkably reduced. 

\smallskip\noindent\textbf{Fine-Tuning Test} Results are shown in~\cref{tab:SK-test}. Compared with RAFT~\cite{teed2020raft}, SEA-RAFT achieves $19.9\%$ improvements on the Sintel clean pass, $4.2\%$ improvements on the Sintel final pass, and $15.7\%$ improvements on KITTI Fl-all score. SEA-RAFT is also competitive among all existing methods in terms of performance-speed trade-off: It is the only method that can achieve results better than RAFT~\cite{teed2020raft} with latency around $70$ms. On Sintel(test), methods with similar performance are at least $1.8\times$ slower than us. On KITTI(test), methods with similar performance are at least $4.6\times$ slower than us.

\subsection{Ablations and Analysis}
\label{sec:ablation}

\setlength\tabcolsep{.2em}
\begin{table*}[t]
    \centering
    \resizebox{1.0\linewidth}{!}{
    \begin{tabular}{lccccccccc}
    \toprule
        \multirow{2}{*}[\multirowcenter]{Experiment} & \multirow{2}{*}[\multirowcenter]{Init.} & \multicolumn{2}{c}{Pre-Training} & \multicolumn{2}{c}{RNN} & \multicolumn{2}{c}{Loss Design} &\multirow{2}{*}[\multirowcenter]{\#MACs} & \multirow{2}{*}[\multirowcenter]{EPE}\\
    \cmidrule(l{0.5ex}r{0.5ex}){3-4}\cmidrule(l{0.5ex}r{0.5ex}){5-6}\cmidrule(l{0.5ex}r{0.5ex}){7-8}
        &  & Img~\cite{deng2009imagenet} & Tar~\cite{wang2020tartanair} & GRU & \#blocks  & Type & Params \\ 
    \midrule
        SEA-RAFT (w/o Tar.) & $\text{\ding{51}}$ & $\text{\ding{51}}$ & - & - & 2 & Mixture-of-Laplace & $\beta_1=0, \beta_2\in[0, 10]$ & 284.7G &0.187\\
    \midrule
         \textcolor{purple}{SEA-RAFT (w/ Tar.)} & $\text{\ding{51}}$ & $\text{\ding{51}}$ & $\text{ \ding{51}}$ & - & 2 & Mixture-of-Laplace & $\beta_1=0, \beta_2\in[0, 10]$ & 284.7G & 0.179\\
    \midrule
         \textcolor{purple}{w/o Img.} & $\text{\ding{51}}$ & - & - & - & 2 & Mixture-of-Laplace & $\beta_1=0, \beta_2\in[0, 10]$ & 284.7G & 0.194\\
    \midrule
        \textcolor{orange}{w/o Direct Reg.} & $\text{\ding{55}}$ & $\text{\ding{51}}$ & - & - & 2 & Mixture-of-Laplace & $\beta_1=0, \beta_2\in[0, 10]$ & 277.3G & 0.201\\
    \midrule
         \textcolor{teal}{RAFT GRU} & $\text{\ding{51}}$ & $\text{\ding{51}}$ & - & $\text{\ding{51}}$ & - & Mixture-of-Laplace & $\beta_1=0, \beta_2\in[0, 10]$ & 297.9G & 0.189\\
    \midrule
         \textcolor{teal}{More ConvNeXt Blocks} & $\text{\ding{51}}$ & $\text{\ding{51}}$ & - & - & 4 & Mixture-of-Laplace & $\beta_1=0, \beta_2\in[0, 10]$ & 314.7G & 0.189\\
    \midrule
         \multirow{2}{*}{\textcolor{blue}{Naive Laplace}} & \multirow{2}{*}{$\text{\ding{51}}$} & \multirow{2}{*}{$\text{\ding{51}}$} & \multirow{2}{*}{-} & \multirow{2}{*}{-} & \multirow{2}{*}{2} & Naive Single Laplace & $\beta\in[-10, 10]$ & \multirow{2}{*}{284.7G}& 0.217\\
            & & & & & & Naive Mixture-of-Laplace & $\beta_1, \beta_2\in[-10, 10]$ & & 0.248\\
    \midrule
        \textcolor{blue}{L1} & $\text{\ding{51}}$ & $\text{\ding{51}}$ & - & - & 2 & $L_1$ & $\text{\ding{55}}$ & 284.7G & 0.206\\
    \midrule
        \textcolor{blue}{Gaussian} & $\text{\ding{51}}$ & $\text{\ding{51}}$ & - & - & 2 & Mixture-of-Gaussian & {$\scalemath{0.8}{\sigma_1=1, \sigma_2=e^{\beta_2}, \beta_2\in[0, 10]}$} & 284.7G & 0.210\\
    \bottomrule
    \end{tabular}
    }
    \caption{We ablate \textcolor{purple}{pretraining}, \textcolor{orange}{direct regression}, \textcolor{teal}{RNN design}, and \textcolor{blue}{loss designs} on Spring~\cite{mehl2023spring} subval. The effect of changes can be identified through comparisons with the first row. See~\cref{sec:ablation} for details.}
    \label{tab:ablations}
    % \vspace{-8mm}
\end{table*}

\begin{figure}[t]
    \centering
    %\vspace{-8mm}
    \includegraphics[width=1.0\linewidth]{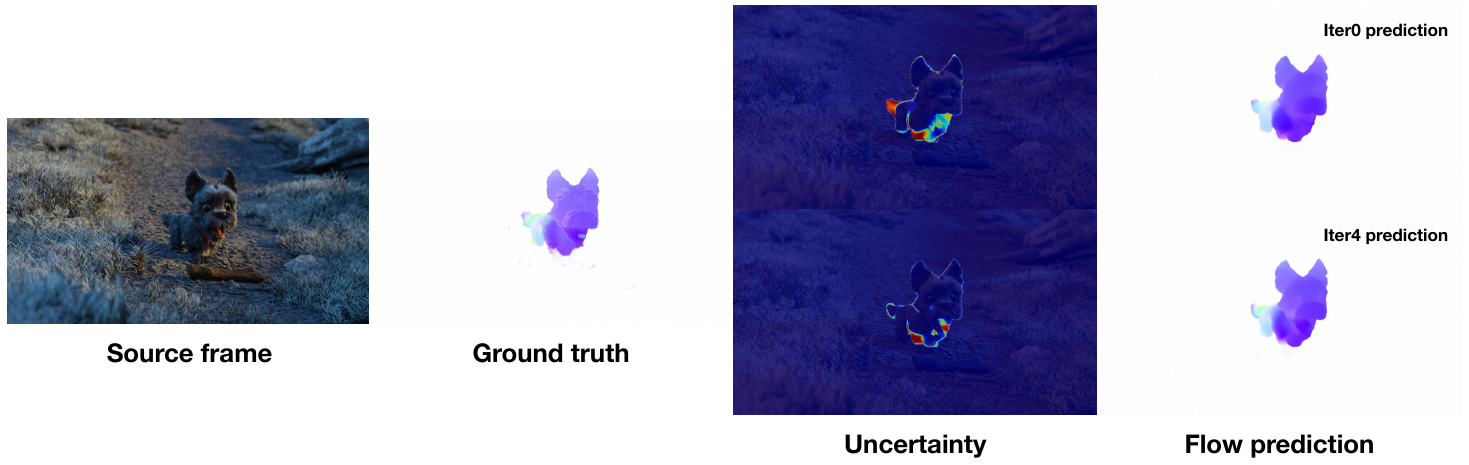}
    \caption{More iterations produce lower variance in the Mixture of Laplace, indicating that the model becomes more confident after each iteration.}
    \label{fig:vis-iter}
\end{figure}

Ablation experiments are conducted on the Spring~\cite{mehl2023spring} dataset based on SEA-RAFT(S). We separate a subval set (sequence 0045 and 0047) from the original training set, train our model on the remaining training data and evaluate the performance on subval. The model is trained with a batch size of 32, input resolution $540\times 960$, and tested following "downsample-upsample" protocol mentioned in~\cref{sec:exp-spring}. We describe the details of ablation studies in the following and show the results in~\cref{tab:ablations}: 

\smallskip\noindent\textbf{\textcolor{purple}{Pretraining}} We test the performance of TartanAir~\cite{wang2020tartanair} rigid-flow pre-training on different datasets(see~\cref{tab:spring,tab:ablations,tab:SK-train} for details). Without TartanAir, SEA-RAFT already provide strong performance, and the rigid-flow pre-training makes it better. We also show that ImageNet pre-trained weights are effective.

\smallskip\noindent\textbf{\textcolor{teal}{RNN Design}} Our new RNN designs can reduce the computation without performance loss compared with the GRU used in RAFT~\cite{teed2020raft}. We also show that on Spring subval, 4 ConvNeXt blocks do not work better than 2 ConvNeXt blocks.

\smallskip\noindent\textbf{\textcolor{blue}{Loss Design}} We see that naive Laplace regression  does worse than the original $L_1$ loss. We also see that it is important to set $\beta_1$ to $0$ in the MoL loss, which aligns the MoL loss to $L_1$ for ordinary cases. Besides, we find that the mixture of Gaussian loss does not work well for optical flow, even though it has been found to be useful for image matching~\cite{chen2022aspanformer}.

\smallskip\noindent\textbf{\textcolor{orange}{Direct Regression of Initial Flow}} We see that the regressed flow initialization significantly improves accuracy without introducing much overhead.

\smallskip\noindent\textbf{Inference Time Breakdown} In~\cref{fig:decomp}, we show how the computational cost increases when we add more refinements. The cost bottleneck for SEA-RAFT is no longer iterative refinements (~\cref{tab:decomp}), which allows us to use larger backbones given the same computational cost constraint as RAFT~\cite{teed2020raft}.

\section{Conclusion}

\setlength\tabcolsep{.4em}
\begin{table}[t]
    \begin{minipage}[b]{.48\linewidth}
        \centering
        \includegraphics[width=1.0\linewidth]{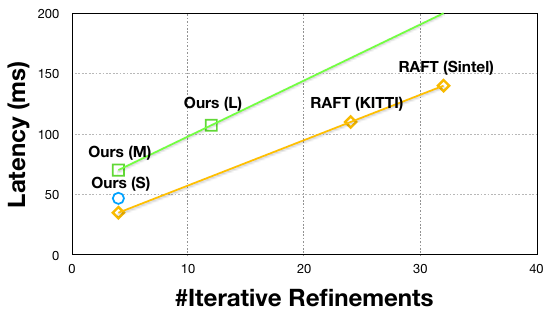}
        \captionof{figure}{Iterative refinements are not hardware-friendly: The latency almost linearly increases with the number of iterations.}
        \label{fig:decomp}
    \end{minipage}\hfill
    {\small
    \begin{minipage}[b]{.48\linewidth}
        \centering
        \begin{tabular}{llcc}
        \toprule
        \multirow{2}{*}[\multirowcenter]{Method} & \multirow{2}{*}[\multirowcenter]{\#Iters} & \multicolumn{2}{c}{Latency (ms)}\\
        \cmidrule(l{0.5ex}r{0.5ex}){3-4}
        & & Total & Iter. \\
        \midrule
            \multirow{2}{*}{RAFT~\cite{teed2020raft}} & 24 (K) & 111 & 90.3 (82\%) \\
            & 32 (S) & 141 & 120 (86\%) \\
        \midrule
            \multirow{3}{*}{SEA-RAFT} & 4 (S) & 47.5 & 18.5 (39\%) \\
            & 4 (M) & 70.9 & 18.5 (26\%) \\
            & 12 (L) & 108 & 55.5 (51\%) \\
        \bottomrule
        \end{tabular}
        \captionof{table}{Compared with RAFT, SEA-RAFT significantly reduces the cost of iterative refinements, which allows larger backbones while still being faster. We use K and S to denote RAFT submissions on KITTI and Sintel respectively.}
        \label{tab:decomp}
    \end{minipage}
    }
\end{table}

We have introduced SEA-RAFT, a simpler, more efficient and accurate variant of RAFT. It achieves high accuracy across a diverse range of datasets, strong cross-dataset generalization, and state-of-the-art accuracy-speed trade-offs, making it useful for real-world high-resolution optical flow.

\section*{Acknowledgements}

This work was partially supported by the National Science Foundation. 
% ---- Bibliography ----
%
% BibTeX users should specify bibliography style 'splncs04'.
% References will then be sorted and formatted in the correct style.
%
\bibliographystyle{splncs04}
\bibliography{main}
\end{document}